# High correlated variables creator machine: Prediction of the compressive strength of concrete

Aydin Shishegaran[a,b], Hessam Varaee[c], Timon Rabczuk[a,d], Gholamreza Shishegaran[e]

[a], Institute of Structural Mechanics, Bauhaus-University Weimar, Marienstr. 15, Weimar D-99423, Germany.

[b], School of Civil Engineering, Iran University of Science and Technology, Tehran, Iran.

[c], Department of Civil Engineering, Ale Taha Institute of Higher Education, Tehran, 14888-36164, Iran

[d], Professor, Department of Geotechnical Engineering, Tongji University, Shanghai, China

[e], School of Civil Engineering, Khajeh Nasir Toosi University of Technology, Tehran, Iran

Abstract.

In this paper, we introduce a novel hybrid model for predicting the compressive strength of concrete using ultrasonic pulse velocity (UPV) and rebound number (RN). First, 516 data from 8 studies of UPV and rebound hammer (RH) tests was collected. Then, high correlated variables creator machine (HVCM) is used to create the new variables that have a better correlation with the output and improve the prediction models. Three single models, including a step-by-step regression (SBSR), gene expression programming (GEP) and an adaptive neuro-fuzzy inference system (ANFIS) as well as three hybrid models, i.e. HCVCM-SBSR, HCVCM-GEP and HCVCM-ANFIS, were employed to predict the compressive strength of concrete. The statistical parameters and error terms such as coefficient of determination, root mean square error (RMSE), normalized mean square error (NMSE), fractional bias, the maximum positive and negative errors, and mean absolute percentage error (MAPE), were computed to evaluate and compare the models. The results show that HCVCM-ANFIS can predict the compressive strength of concrete better than all other models. HCVCM improves the accuracy of ANFIS by 5% in the coefficient of determination, 10% in RMSE, 3% in NMSE, 20% in MAPE, and 7% in the maximum negative error.

1. Introduction

Thanks to its low cost, concrete is among the most widely used building materials in the world. Estimating the properties of hardened concrete has been the topic of numerous studies [1]. The compressive strength is a key property of concrete and directly related to the safety of the structure. Evaluating the concrete compressive strength (CCS) is a common requirement in quality control and performance determination of existing structures during their entire life cycle. Strength test results also may be a necessity after a structural failure, such as environmental degradation or fire damage [2].

Methods to estimate the mechanical properties of concrete can be classified into destructive tests (DTs) and non-destructive tests (NDTs). Usually, cubic or cylinder specimen are prepared and cured to measure the compressive strength of fresh concrete in accordance with procedures described in ASTM standard test methods [3,4]. Different concrete mixtures lead to different properties, and reproducibility of the test results is of major importance [5–8]. Although DTs, such as core tests, are commonly used, they have several drawbacks. They are expensive, time-consuming and have 'locally destructive effects' on the structures. They are also only representative for small structures and conducting these tests are sometimes difficult due to accessing of the coring machine [9].

NDTs are widely used to evaluate the fundamental characteristics of concrete in structures, such as compressive strength. NDTs do not cause any damage to the structure and do not affect the structural performance [10]. Many different types of NDTs have been developed in order to measure the compressive strength of concrete, among them the ultrasonic pulse velocity (UPV) and rebound hammer (RH) tests. These methods are simple yet effective and can be used for quality control of concrete in structures or confirm the uniformity of the material from one part to another [11–13].

UPV is a common and fast test without any damage on concrete. It was presented and proposed by ASTM C 597 and BS 1881: Part 203 [4,14]. This test is one of the suitable approaches to check the uniformity of concrete quickly, and it can be applied in various parts of reinforced concrete (RC) elements. This test can determine the properties of used concrete in structures and also specify the presence of voids and internal cracks [15]. The results of this test depend on several factors, including the percentage of components of the mix proportion, properties of aggregates, steel bars, cracks, voids and fibers. On the other hand, some studies report high uncertainties of the results of UPV to estimate the compressive strength of concrete [16,17].

The RH test is one of the most used non-destructive tests because of its simplicity. This test was presented by ASTM C 805, and BS 1881: Part 202 [14,18]. In this test method, a spring-loaded mass, which has a fixed amount of energy, is released. The traveled distance by this mass is defined as a percentage of the initial spring extension. This value is defined through a rebound number (RN). The determined RNs depend on the percentage of components of mix proportions, surface texture, surface wetness of the concrete and properties of the aggregate, although the RN is the reflection of the concrete surface [11,19,20].

Sometimes, a combination of UPV and RH methods is used for better prediction of concrete compressive strength, which is also known as the SonReb method. UPV and RH tests are both sensitive to the variations in the concrete properties, especially in opposite directions. UPV tests provide information about the interior concrete properties whereas the RH test results usually predict concrete strength near the surface. Consequently, using the combination of two methods for the prediction can lead to more accurate and reliable results [19,21].

2. Theoretical background

Introducing precise and reliable models for predicting the CCS would lead to saving both costs and time. In this regard, many researchers have used several models to estimate the compressive strength of concrete [22–27]. Based on the results of the previous studies, computational modeling, artificial intelligence (AI), and parametric multi-variable regression models are three groups of techniques that are commonly used to predict concrete strength using the UPV and RH outputs. Computational modeling is based on the mathematical modeling of complex physical phenomena and often encounter some difficulties due to the complexity of the models or the time consuming numerical computation [28]. In this regard, several equations were established by researchers for predicting concrete compressive strength; however, most of them require previous knowledge of concrete ingredients to achieve reliable results. Among them, Rashid and Waqas [11] established a relationship between UPV with compressive strength and RN with compressive strength. Ahmadi-Nedushan proposes an optimized instance-based learning approach for prediction of the compressive strength of high performance concrete based on mix data, such as water to binder ratio, water content, super-plasticizer content, fly ash content [27]. In recent years, AI methods and machine learning (ML) techniques have been used frequently to predict the CCS without knowing the theoretical relationships between the input and the output. Several researchers have employed AI methods to predict the mechanical properties of different types of concrete using detail of mix proportion and UPV [24,24,28–30]. Tenza-Abril et al. used an artificial neural network (ANN) for prediction and sensitivity analysis of compressive strength in segregated lightweight concrete using UPV [31]. Ashrafian [32] utilized heuristic regression methods as a hybrid method in the prediction of both strength and UPV for fiber reinforced concrete. Adaptive neuro-fuzzy inference system (ANFIS) is also a strong model for predicting the performance of concrete. For example, the properties of rubberized concrete, compressive strength and splitting tensile strength of concrete containing silica fume and recycled aggregate at 3, 7, 14, 28, and 90 days, and compressive strength of no-slump concrete were predicted using ANFIS [23,33,34]. Cevik (2011) employed ANFIS to predict the increased strength of fiber-reinforced polymer (FRP) confined concrete [35]. Several researchers used several methods, including ANFIS, ANN, and group method, to estimate the compressive strength of concrete based on the results of core tests. The used variables include length-to-diameter ratio, core diameter, aggregate size, and concrete age [36].

Parametric multi-variable regression models are another group of methods to predict CCS. Specifically, the confidence interval of a prediction calculated by a parametric regression model reveals the accuracy of the estimation, and the statistics of the model parameters can indicate how each variable in the model influences the prediction [37]. Kewalramani [16], used a regression analysis to predict the compressive strength of concrete based on acoustic characteristics like UPV and damping constant.

The main target of this study is a comparison of the accuracy and performance of two new proposed models and two available models in predicting the compressive strength of concrete using RN and UPV. First, the data is collected from 8 previous studies. Second, the step by step regression (SBSR), GEP, and ANFIS are used as an individual model to predict the compressive strength. SBSR model is introduced and used for the first time. Third, high correlated variable creator machine (HCVCM) is used to generate the new variables, which can improve the accuracy and performance of the models. HCVCM is also introduced and utilized for the first time. Using the combination of HCVCM and the used models in previous steps, three hybrid models, including HCVCM-SBSR, HCVCM-GEP, and HCVCM-ANFIS, are created and used to predict the compressive strength. The statistical parameters and error terms, including coefficient of determination, root mean square error (RMSE), normalized square error (NMSE), fractional bias, the maximum positive and negative errors, and mean absolute percentage error (MAPE), are used to evaluate and compare the models. The statistical parameters and error terms are used to compare the models. It is clear that using HCVCM and SBSR is the novelty of this study.

3. Materials and methods

The main target of this study is related to the comparison accuracy and performance of two unknown models and two known models in predicting the compressive strength of concrete using RN and UPV. First, the values of RN, UPV and compressive strength of concrete from 8 previous studies are collected. Then, an unknown regression model namely SBSR, GEP, and ANIFS are utilized to predict the compressive strength of concrete using the initial input variables, including the values of RN and UPV. Subsequently, HCVCM is used to propose several new variables. The proposed new variables by HCVCM are imported as input variables in SBSR, GEP, and ANFIS; therefore, HCVCM-SBSR, HCVCM-GEP, and HCVCM-ANFIS can be regarded as hybrid models. The results of single models are compared with the results of the hybrid models. 70% of the dataset, which is selected to train models randomly, is the same for all models. After training the models, the remaining dataset is predicted using the trained models.

3.1. Data

To build up a precise prediction model, a large number of experimental data on normal-weight concrete compressive strength, UPV test results, and rebound number from rebound hammer tests is required. We have collected 516 data points of the compressive strength of concrete, which were tested at 3, 7, 14, 28, 56, and 90 days [9,10,13,15,17,38–40].

(a)

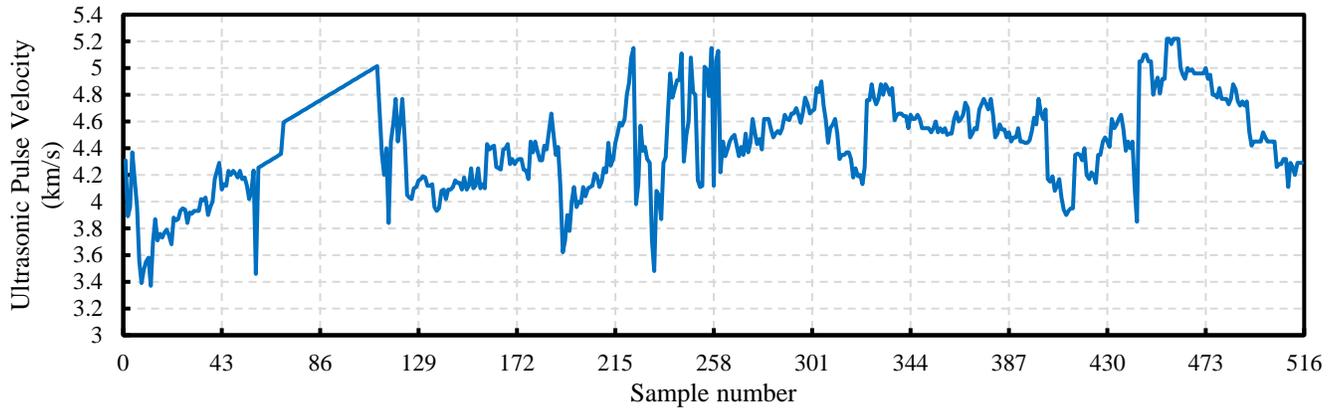

(b)

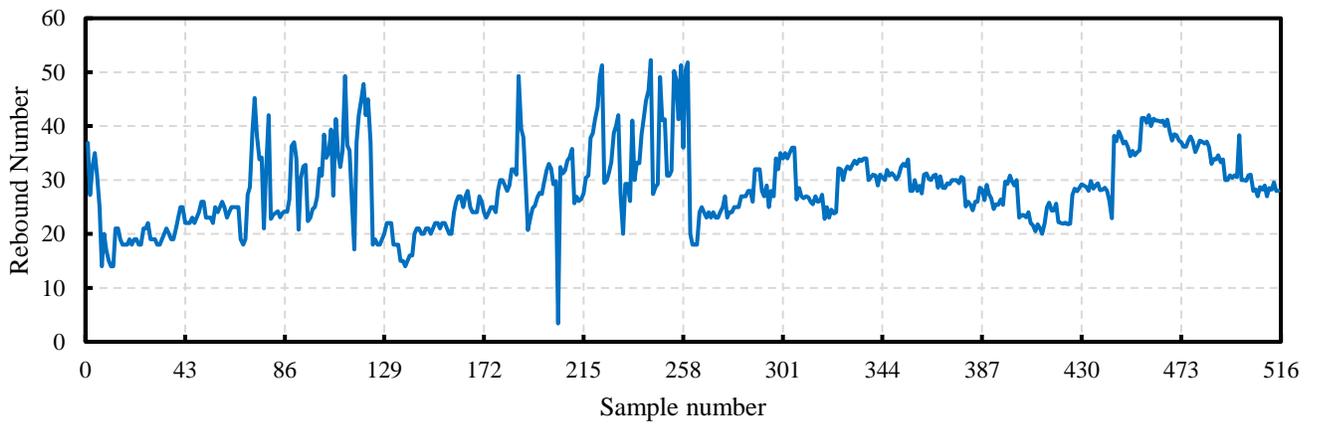

(c)

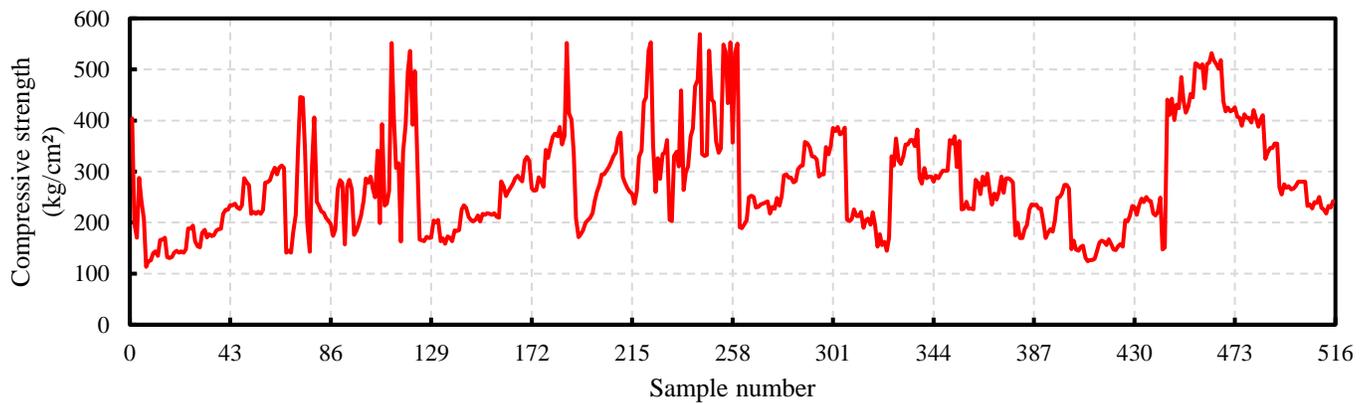

Fig 1. The values of input variables and output. (a) UPV. (b) RN. (c) Compressive strength of concrete

UPV and RN are considered as inputs of the models, and the compressive strength of concrete is selected as the output. Table 1 shows the variation range, average, median, and standard deviation of each variable, while the statistical distribution of the involved variables can be found in Fig. 2.

**Table 1** Statistical parameters of input variables and output

| Variable | Unit | min | max | average | SD | median | Type |
|---|---|---|---|---|---|---|---|
| UPV | Km/S | 3.37 | 5.22 | 4.44 | 0.35 | 4.45 | Input |
| RN | - | 3 | 52 | 29 | 7.32 | 28 | Input |
| CCS | Kg/Cm$^2$ | 113.33 | 569.21 | 276.92 | 98.60 | 261.97 | Output |

(a)

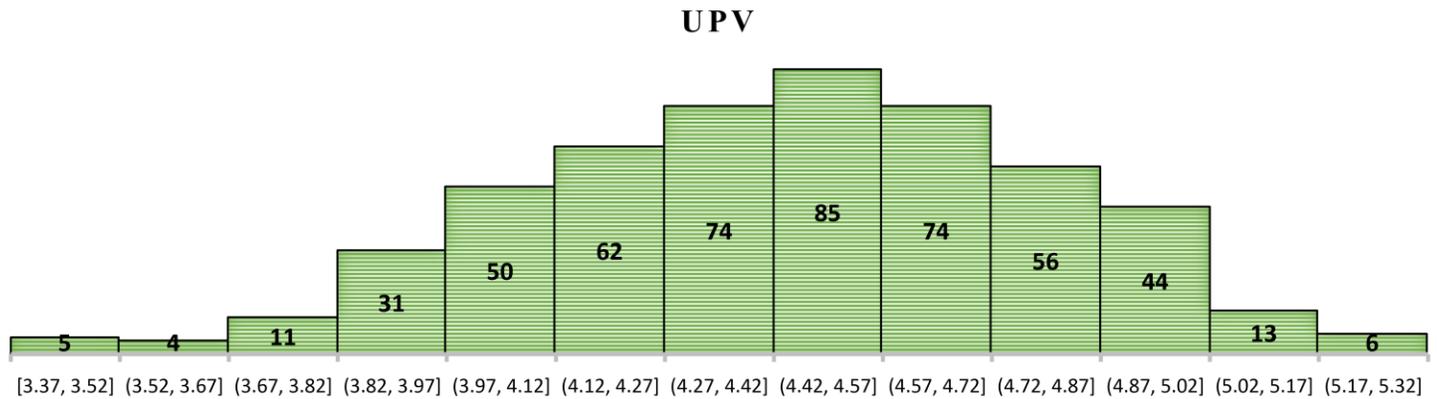

(b)

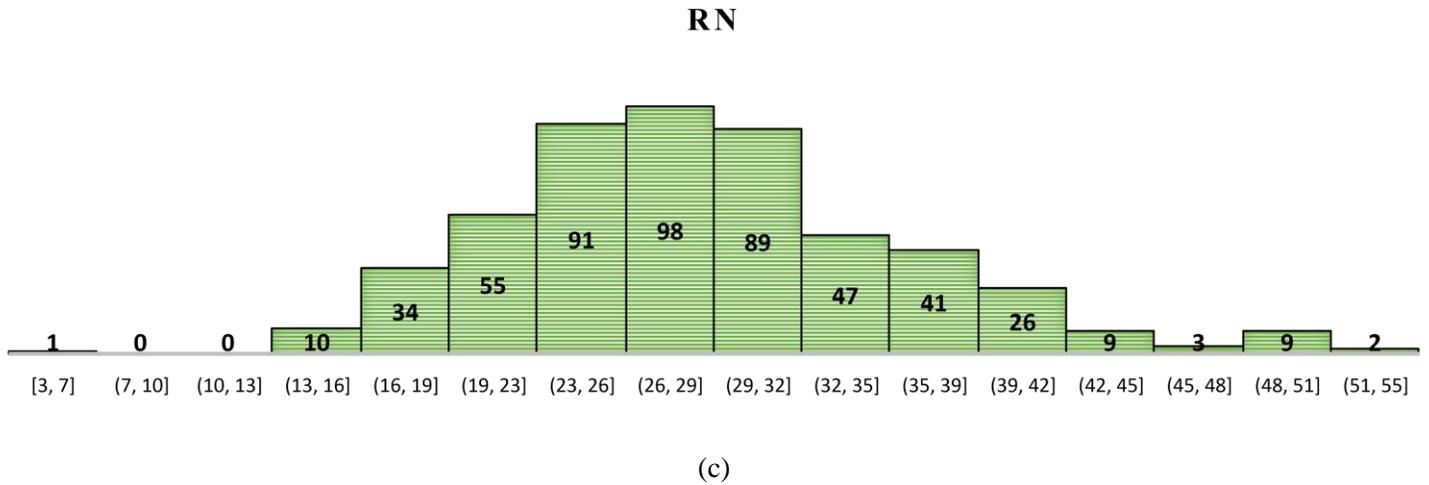

(c)

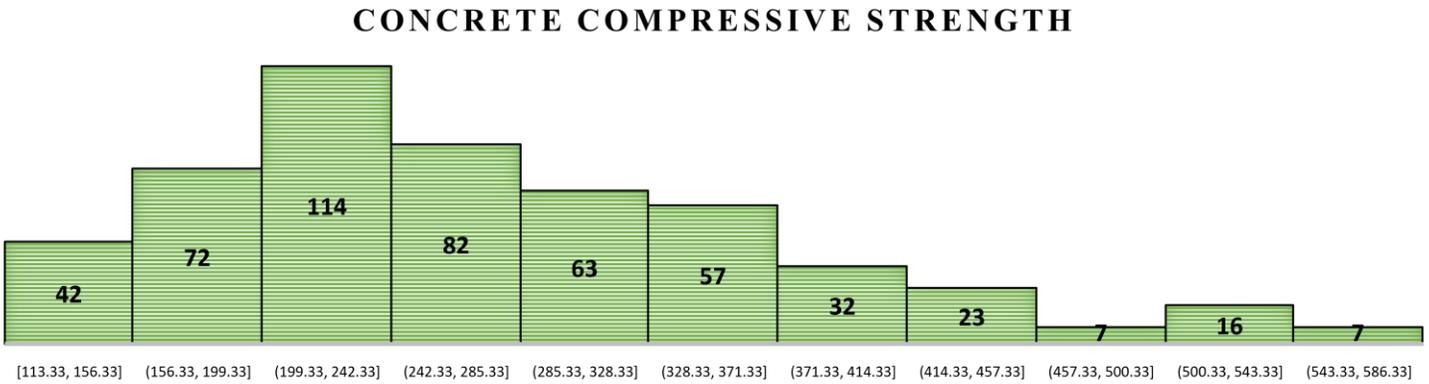

Fig. 2. The statistical distribution of input variables and output. (a)UPV. (b)RN. (c) compressive strength

Table 2 shows the coefficient of determination between the input variables and also the coefficient of determination between the inputs and output. The coefficient of determination is a statistical measurement that examines how differences in one variable can be explained by the difference in a second variable [41,42]. In other words, this coefficient assesses how strong the linear relationship is between two parameters. The optimal value of the coefficient of determination between an input and the output is 1. Several input variables might have the same coefficient of determination with the output. In these cases, if the coefficient of determination between these input variables is close to 1, it means that the input variables represent the same properties and data. Table 2 shows the values of the coefficient of determination between each input and output and between input variables. The results show that the input variables are independent, and RN has a better correlation with the compressive strength of concrete.

**Table 2.** The coefficient of determination between variables and output

|  | Coefficient of determination |
|---|---|
| Between compressive strength and UPV | 0.443 |
| Between compressive strength and RN | 0.758 |
| Between RN and UPV | 0.513 |

3.2. Methods

In this study, three single algorithms and three hybrid algorithms are used to predict the compressive strength of concrete using RN and PUV. The single algorithms include SBSR, GEP and ANFIS. The hybrid models are created by combining HCVCM and each single algorithm. 70% of the dataset is selected randomly to train the models. Then, they are used to predict the remaining dataset.

3.2.1. Step by step regression (SBSR)

Step by Step Regression (SBSR)

SBSR is a new regression model, which focuses on coefficient of determination between variables and output to improve the accuracy of the obtained model. Fig 3. shows a SBSR model for three variables. In the first step, the coefficient of determination between each input variable and output is calculated. Then, the variables are sorted based on the obtained values of coefficient of determination. The variable with the highest coefficient of determination is imported in cell 1, and the variables with the smallest coefficient of determination is placed in the last cell. In the second step, a linear regression is used to predict the output in each cell. As shown in Fig 3, the predicted output of each cell is imported as a new variable in the next cell. In the third step, the predicted outputs in all cells are imported as input variables to predict the actual output. In all cells, the constant parameters of the regression model are obtained using the mean square error (MSE) technique. Equations (1)-(3) show the used regression model in the first cell, middle cells, and the final cells, respectively. Equation (4) represent the regression model in the final step.

$$Y = b_1 + b_2 \times X_1 \tag{1}$$

$$Y = b_3 + b_4 \times X_2 + b_5 \times F_1 \tag{2}$$

$$Y = b_6 + b_7 \times X_3 + b_8 \times F_2 \tag{3}$$

$$Y = b_9 + b_{10} \times F_1 + b_{11} \times F_2 \tag{4}$$

where $X_1$, $X_2$, and $X_3$ are defined as the sorted input variables based on the coefficient of determination with the output; $F_1$ and $F_2$ are defined as the outputs of Cells 1 and 2, respectively while $b_1$, $b_2$, …, $b_{11}$ denote the constant variable of the regression model, i.e.

$$b_i = (\frac{X^T}{X})^{-1}(\frac{X^T}{Y}) \tag{4}$$

where the superimposed 'T' refers to the transpose.

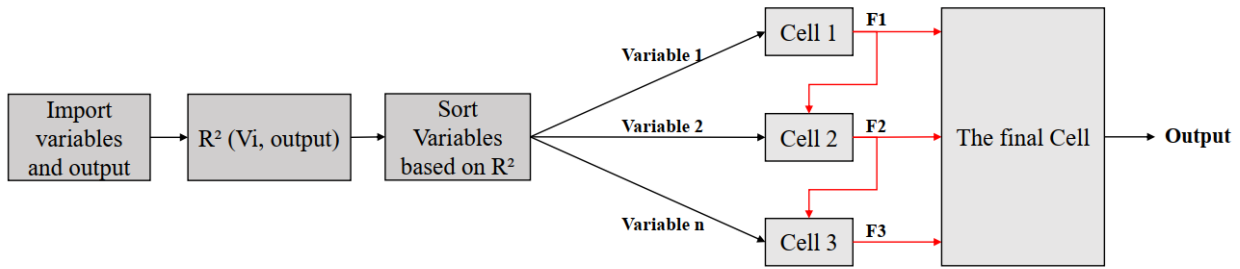

Fig 3. The general form of SBSR

### 3.2.2. Gene expression programming (GEP)

GEP has been introduced by [43]. It consists of Genetic Algorithm (GA) and Genetic Programming (GP) [43] and is a strong tool to predict various phenomena and material properties. This model is created by using a set of different functions and terminals. A set of different functions, which are used in this study, includes the basic functions, inverse trigonometric function, trigonometric functions, exponential function, hyperbolic function, user-defined functions and a combination of them. A set of terminals consists of input variables, their combination and constant values. In this approach, a random population of data is used by GA to find a suitable function to predict the output, which is the compressive strength of concrete in this study. GA selects the suitable input variables and functions between all functions and input variables to estimate the output. Several operators, named genes, are used for performing Genetic variations. Based on the GEP algorithm, the Roulette wheel selects data. Several genes reproduce the data. In the duplication process, the genes avoid unsuitable data and transfer the suitable data to create the next generation. The main target of the mutation operator is related to the internal random optimization of the given chromosomes. Fig 4 explains the process of GEP.

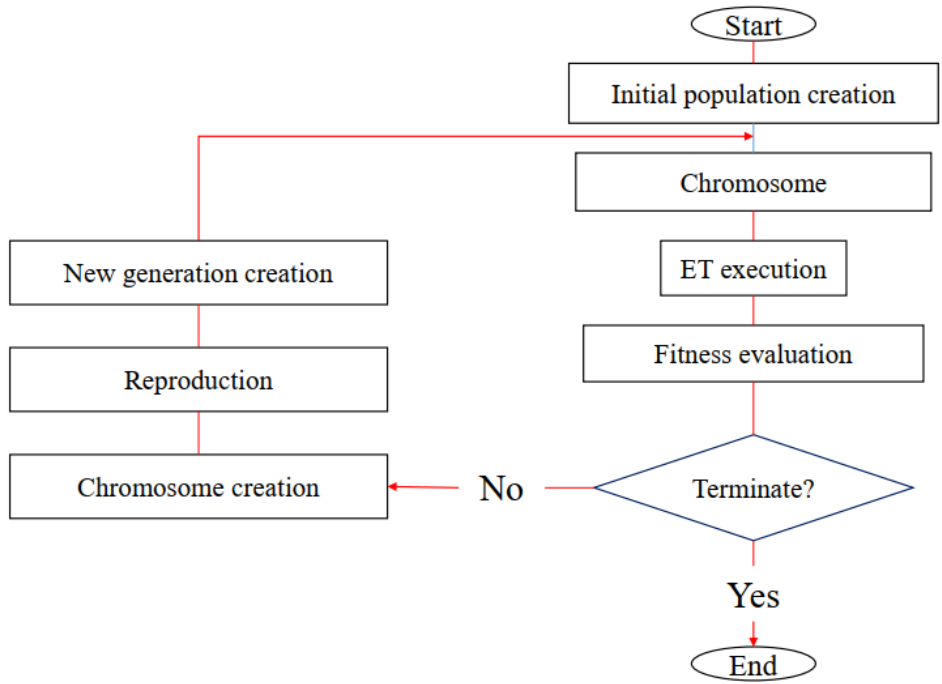

Fig 4. The process of GEP

### 3.2.3. Adaptive neuro-fuzzy inference system (ANFIS)

Fuzzy system is a rule-based system, which is created by selecting linguistic rules. This system has suitable accuracy in predicting material performance. Two kinds of fuzzy inference system (FIS) were used in most applications, i.e. Sugeno fuzzy and Mamdani models. Their defuzzification and aggregation approaches are different; therefore, the results of fuzzy rules are various for these systems. In the present investigation, the rule is defined as a linear combination of input parameters, and the final output equals to the weighted mean of outputs for each rule. For example, A Sugeno FIS, which includes an output (f), and two input parameters (y and x), is represented as follows:

$$\text{Rule 1: if } x \text{ is } A_1 \text{ and } y \text{ is } B_1 \text{ then } f_1 = p_1 x + q_1 x + r_1 \tag{5}$$

$$\text{Rule 2: if } x \text{ is } A_2 \text{ and } y \text{ is } B_2 \text{ then } f_2 = p_2 x + q_2 x + r_2 \tag{6}$$

where $B_1$, $B_2$, $A_1$, and $A_2$ are the labels presenting the membership functions (MFs) for the inputs x and y, respectively; $r_i$, $q_i$, and $p_i$ designate as the output membership function parameters (OMFPs), which are the consequent parameters of *i*th rule.

The performance and accuracy of the fuzzy systems are related to the used rule based and the parameters such as the MFs, and they are tuned and adjusted using the combination of ANN with fuzzy system to create neuro-fuzzy system. On the other hand, a neuro-fuzzy system includes the learning processes in ANN and the fuzzy systems; thus, this model has the advantage of both models. ANFIS is one of these combinations, which uses an adaptive back propagation learning procedure for adjusting the parameters in the fuzzy systems [44]. In other words, ANFIS utilizes a linguistic model containing a set of If-Then fuzzy rules and the fuzzy sets to present the human-like reasoning style of a fuzzy system [45]. Fig 5 (a) demonstrates the general architecture of the ANFIS model, which include two input parameters, and Fig 5 (b) shows the fuzzy reasoning mechanism.

The rule base of ANFIS, including two Fuzzy If-Then rules of Sugeno and Takagi, is assumed as below [46,47]:

Rule 1: if x is $A_1$ and y is $B_1$ Then $f_1 = p_1x + q_1x + r_1$   (7)

Rule 2: if x is $A_2$ and y is $B_2$ Then $f_2 = p_2x + q_2x + r_2$   (8)

Each layers' functions are explained as follows:

The first Layer: Each node (i) in the first layer is a square node with a node function:

$$O_A^i = \mu_{Ai}(x) \qquad (9)$$

where O denotes the output of each node in this layer; x and $A_i$ are the input variable to node (i) and the label to present the MFs for the input, respectively. The Gaussian membership functions (GMFs), used here are given by

$$\mu_{Ai}(\ln_1, \sigma_i, C_i) = \exp(-\frac{(\ln_1 - C_i)^2}{2\sigma_i^2}) \ , \ i = 1, 2 \qquad (10)$$

$$\mu_{Bj}(\ln_2, \sigma_j, C_j) = \exp(-\frac{(\ln_1 - C_j)^2}{2\sigma_j^2}) \ , \ j = 1, 2 \qquad (11)$$

$\{\sigma_i, C_i\}$ and $\{\sigma_j, C_j\}$ being the premise parameters, which are governed form GMFs.

The second layer: Each node in the second layer is a circle node, which is labeled as N that multiplies the output signals, and their outputs are calculated as follows:

$$W_i = \mu_{Ai}(x)\mu_{Bi}(y) \ \ i = 1, 2 \qquad (12)$$

In this layer, the outputs of the nodes show the firing weight of the rules.

The third layer: Each node in the third layer is defined as a circle node, which is labeled as N that plays a normalization role. The *i*th node calculates the ratio of the firing weight of the *i*th rule to the summation of the firing weights of all rules as:

$$\overline{W}_i = \frac{W_i}{W_1 + W_2} \quad i = 1, 2 \tag{13}$$

The fourth layer: Each node in the fourth layer is a square node with a node function. Each output of this layer is determined as:

$$O_i^4 = \overline{W}_i f_i = \overline{W}_i (p_i x + q_i y + r_i) \quad i = 1, 2 \tag{14}$$

The parameters of layer 4 are related to as consequent parameters.

The fifth layer: There is one circle node in layer 5, which is a fixed node labeled by $\Sigma$, and defined as the signal node. This node computes the overall output, which is obtained from the summation of all incoming signals as

$$O_i^5 = \Sigma \overline{W}_i f_i = \frac{\Sigma W_i f_i}{\Sigma W_i} \tag{15}$$

There are two approaches in ANFIS. The first is a back propagation for all parameters. The second is a hybrid approach, which includes a least squares estimation for the parameters associated with the OMFPs and a back propagation for the parameters associated with the input MFs [46].

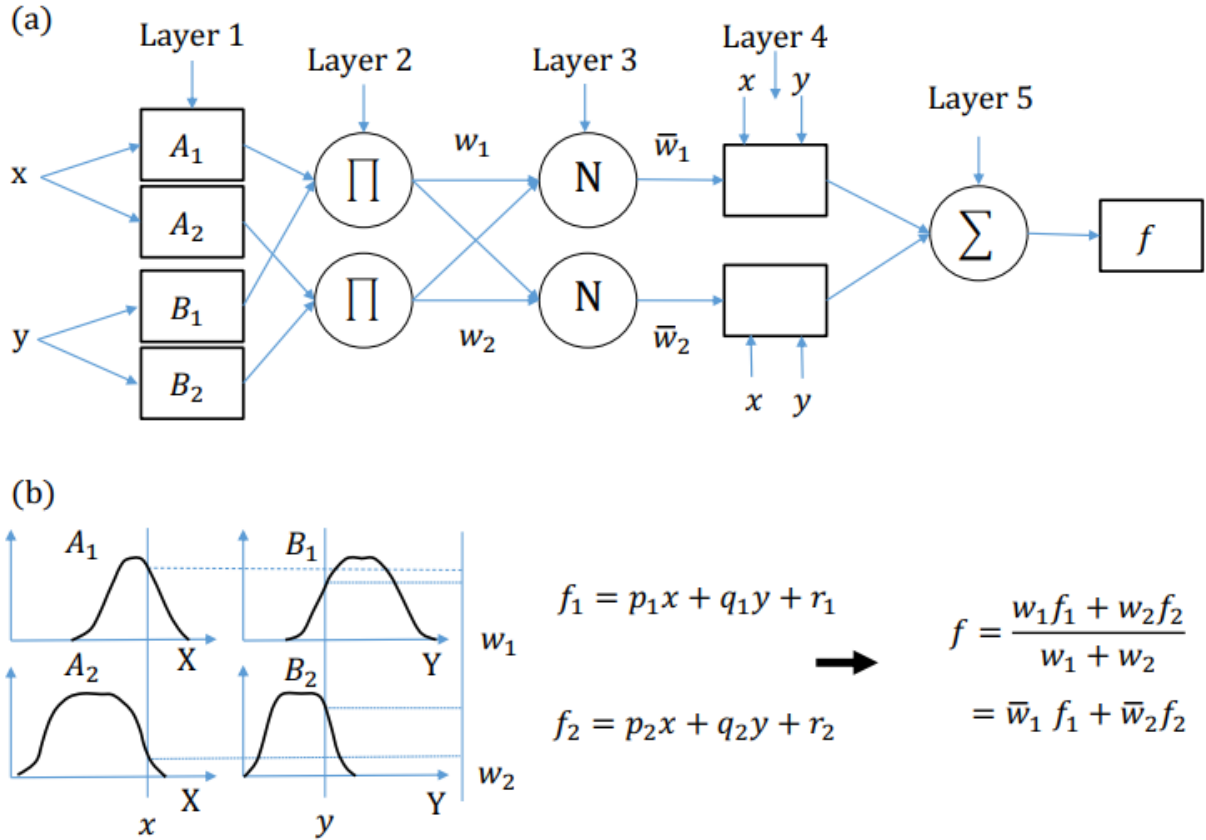

Fig 5. The structure of ANFIS (a) The general architecture of ANFIS (b) Fuzzy reasoning scheme of ANFIS [46]

3.2.4. High correlated variables creator machine (HCVCM)

HCVCM is a new algorithm to create the new variables from the initial variables improving the performance of model. There are three steps in this method. In the first step, the new variables are created using several mathematical, trigonometric functions, their combinations and user-defined functions. In the second step, the coefficient of determination between new variables and outputs is determined. The new variables with highest coefficient of determination with the output are imported to the third steps. In the third step, the new variables are selected and imported to the prediction models with the lowest coefficient of determination between the initial variables. HCVCM can contain several generations. Fig 6. shows the process of the first generation in HCVCM. Only new variables, which are selected in the first generation, are imported as input variables in our hybrid models. In the second generation, new variables can be created using the new selected variables in generation 1.

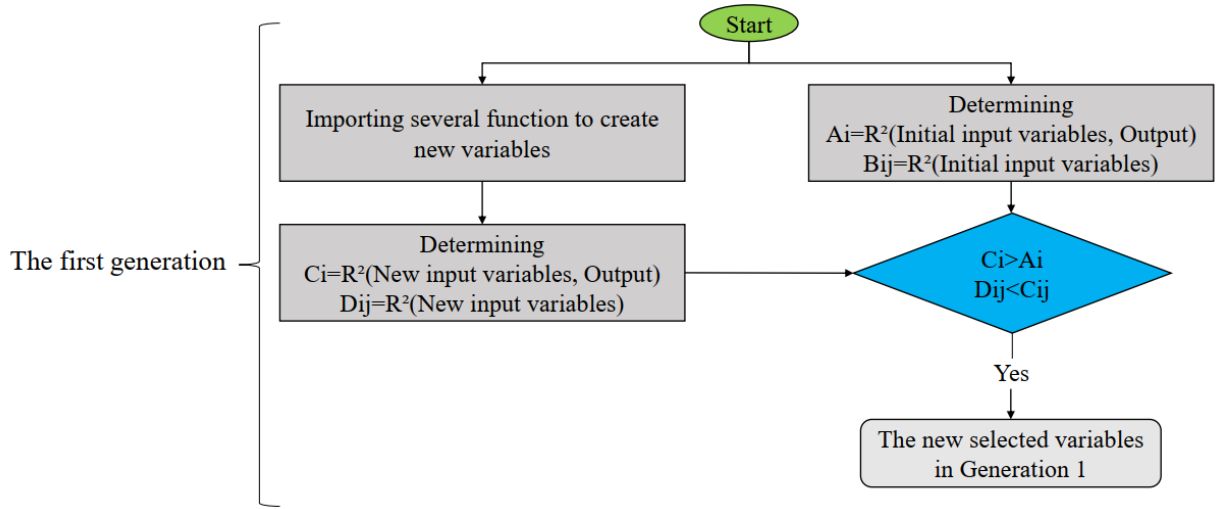

Fig 6. The process in HCVCM

3.2.5. Using HCVCM in prediction models

HCVCM is used before SBSR, GEP, and ANFIS to create the new variables, which finally improves the accuracy of the models. First, better variables are specified based on HCVCM. Then, these variables are imported to SBSR, GEP, and ANFIS. Based on the results of HCVCM, the $R^2$ value of 6 new variables created form UPV will be higher compared to the f $R^2$ value of UPV. Moreover, the $R^2$ value of the new RN will be higher than the old $R^2$ value. The functions, which are used to create these new variables from UPV and RN, are shown in table 3.

Table 3. The coefficient of determination between new variables and output

| New variables | UPV2 | UPV5 | UPV6 | UPV7 | UPV8 | UPV9 | RN5 |
|---|---|---|---|---|---|---|---|
| | Cos(a) | $a^2$ | $a^3$ | $a^4$ | $a^5$ | Exp(a) | $b^2$ |
| Coefficient of determination (new variable, output) | 0.454 | 0.455 | 0.464 | 0.471 | 0.476 | 0.478 | 0.776 |

As mentioned in section 3.2.4, not only the coefficient of determination between the new variables and the output should be higher than the coefficient of determination between the initial variable and the output, but also the coefficient of determination between the new variables should be lower than the coefficient of determination between the initial variables. Therefore, as shown in tables 3 and 4, the new variables are selected, which pass these two rules.

Table 4. The coefficient of determination between new variables

| New variables | UPV2 | UPV4 | UPV5 | UPV6 | UPV7 | UPV8 |
|---|---|---|---|---|---|---|
| | RN5 | RN5 | RN5 | RN5 | RN5 | RN5 |
| Coefficient of determination | 0.484 | 0.434 | 0.481 | 0.493 | 0.502 | 0.509 |

4. Results

As mentioned before, 70% of the dataset was used to train the models, then the model was utilized to predict the remaining. The same data was used to train all models. Table 5 summarizes the statistical parameters of six models in the training and testing dataset. ANFIS performs better than all the other single models with respect to all statistical parameters in the training dataset, and SBSR is better than GEP according to the coefficient of determination, RMSE, and NMSE. In contrast, GEP performs better than the other single models in the testing dataset. Although the coefficient of determination of SBSR is better than those of ANFIS in the testing dataset, ANFIS shows better results with respect to RMSE and NMSE. The results show that the best single model based on statistical parameter cannot be selected.

As table 5 demonstrates, HCVCM positively affects the results of SBSR and ANFIS because not only the $R^2$ value increases – compared to the associated single models, but also the RMSE and NMSE are less. HCVCM-GEP performs better than GEP in the training dataset, although in the testing dataset, GEP yields better results in the RMSE and coefficient of determination.

Table 5. The results of the statistical parameters for all models

| Models | Coefficient of determination | RMSE | NMSE | Fractional bias | Coefficient of determination | RMSE | NMSE | Fractional bias |
|---|---|---|---|---|---|---|---|---|
| SBSR | 0.761 | 46.813 | 0.084 | -1.56E-05 | 0.935 | 51.119 | 0.189 | 7.97E-05 |
| HCVCM-SBSR | 0.788 | 43.809 | 0.100 | -1.41E-05 | 0.926 | 43.538 | 0.130 | 7.20E-05 |
| GEP | 0.756 | 47.824 | 0.105 | -2.54E-05 | 0.943 | 36.979 | 0.109 | 7.22E-05 |
| HCVCM-GEP | 0.767 | 46.207 | 0.102 | -1.41E-05 | 0.911 | 39.273 | 0.108 | 4.36E-05 |
| ANFIS | 0.824 | 39.901 | 0.090 | -1.66E-05 | 0.930 | 39.297 | 0.110 | 8.24E-05 |
| HCVCM-ANFIS | 0.862 | 35.742 | 0.064 | -1.35E-05 | 0.950 | 32.628 | 0.084 | 8.03E-05 |

Table 6 lists the error terms including the maximum positive error, maximum negative error and MAPE. ANFIS is the best single model. HCVCM-ANFIS performs better than ANFIS with respect to the error term. It also leads to better results compared to HCVCM-SBSR and HCVCM-GEP. The maximum positive and negative errors of HCVCM-SBSR are less than those of HCVCM-GEP, but MAPE of HCVCM-GEP is less than those of HCVCM-SBSR. The results show that HCVCM is more useful when this method is used with SBSR and ANFIS.

Table 6. The results of the error terms for all models

| Error terms | SBSR | HCVCM-SBSR | GEP | HCVCM-GEP | ANFIS | HCVCM-ANFIS |
|---|---|---|---|---|---|---|
| The maximum positive error | 102.3 | 31.732 | 35.813 | 39.175 | 28.842 | 31.706 |
| The maximum negative error | -89.7 | -81.815 | -86.275 | -86.320 | -85.303 | -79.363 |
| MAPE | 15.73 | 13.932 | 13.658 | 12.856 | 12.174 | 10.145 |

Fig 7 (a) shows the comparison of the actual compressive strength and the predicted values using SBSR and HCVCM-SBSR. The HCVCM-SBSR results better predict the actual compressive strength. Fig 7(b) depicts the error of these models. HCVCM decreases the error in SBSR. Figs. 8 (c) and (d) compare the predicted compressive strength by GEP and HCVCM-GEP with the actual compressive strength. HCVCM is obviously not useful for GEP because GEP was not used the new variables, which were proposed by HCVCM. On the other hand, GEP changed all new variables, which were created by HCVCM; therefore, the results of HCVCM-GEP cannot improve the results of GEP. Fig 7 (e) shows the predicted compressive strength using ANFIS and HCVCM-ANFIS whereas Fig. 7 (f) illustrates the obtained errors from ANFIS and HCVCM-ANFIS.

(a)                                                                                          (b)

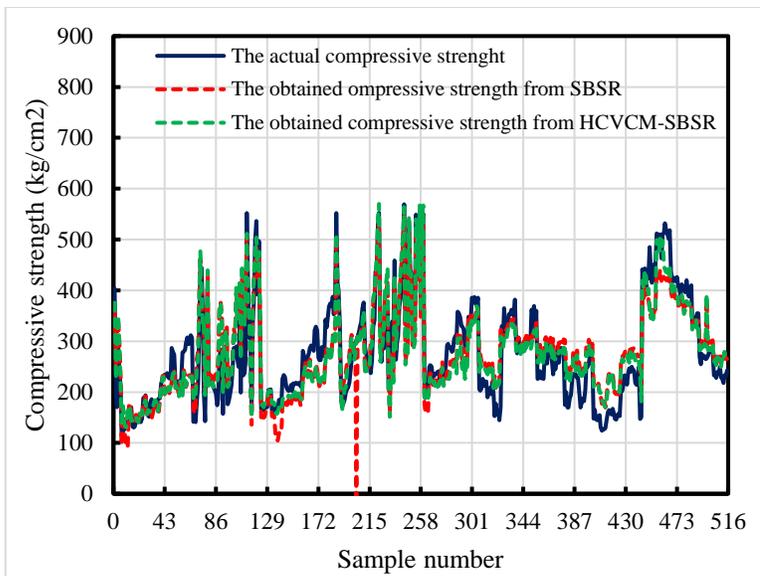

(c)

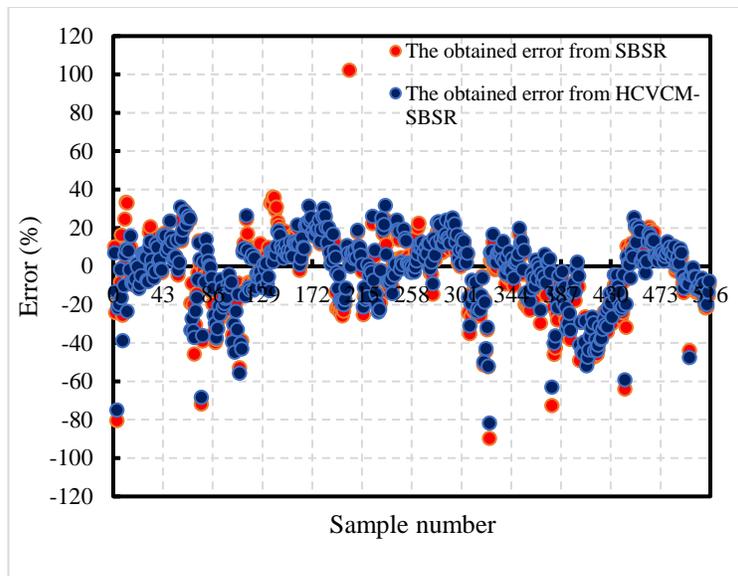

(d)

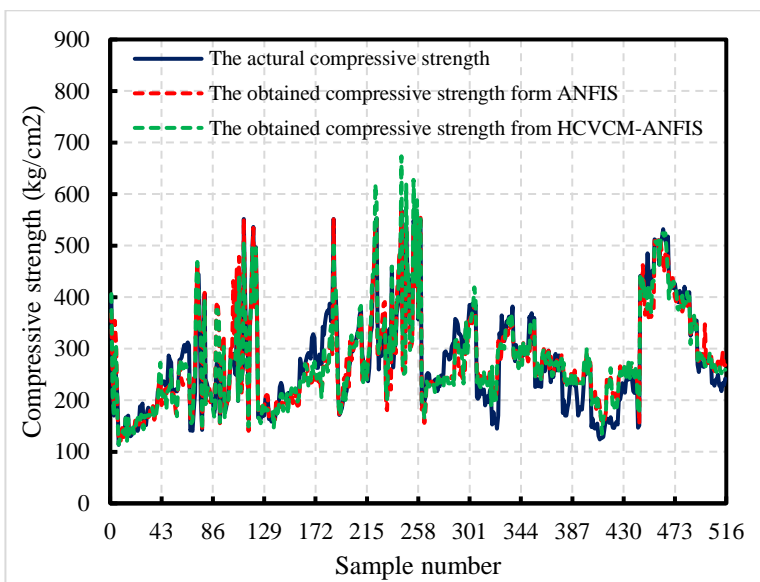

(e)

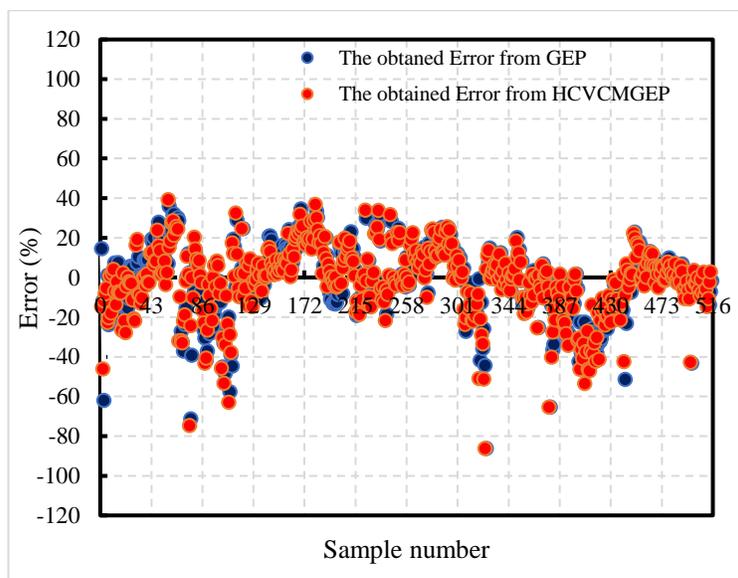

(f)

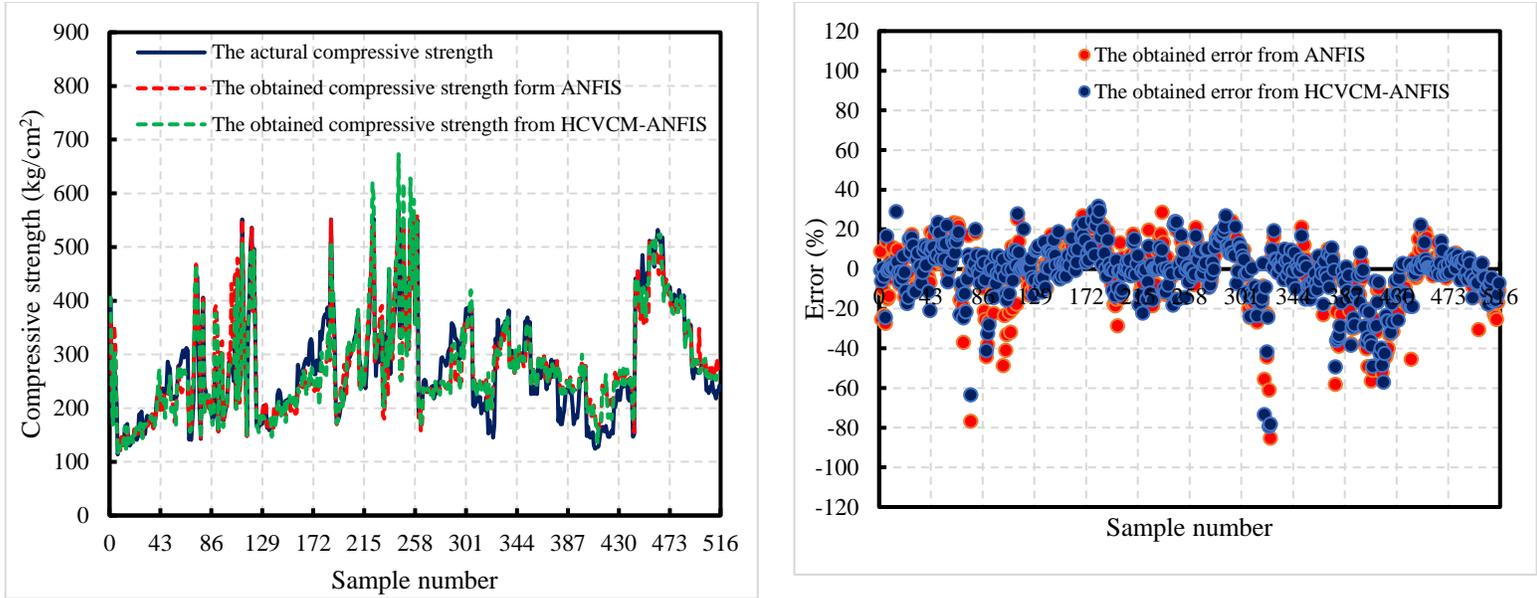

Fig 7. The actual and predicted compressive strength of concrete and obtained errors from models. (a) The predicted compressive strength by SBSR and HCVCM-SBSR. (b) The obtained error from SBSR and HCVCM-SBSR. (c) The predicted compressive strength by GEP and HCVCM-GEP. (d) The obtained error from GEP and HCVCM-GEP. (e) The predicted compressive strength by ANFIS and HCVCM-ANFIS. (f) The obtained error from ANFIS and HCVCM-ANFIS.

5. Conclusion

In this study, three single models, including SBSR, GEP, and ANFIS, and three hybrid models, including HCVCM-SBSR, HCVCM-GEP, and HCVCM-ANFIS, were used to predict the compressive strength of 516 concrete samples with various UPV and RN. In the following, the main findings are summarized:

1- HCVCM aims to produce better variables with respect to the coefficient of determination between the new variables and the output. On the other hand, the coefficient of determination between new input variables and outputs is more than those between the initial variables and output.

2- Selecting the new variables with lowest correlation to each other and highest correlation with the output can improve the accuracy of the models.

3- HCVCM improves the accuracy of SBSR and ANFIS to predict the compressive strength of concrete.

4- HCVCM is an adaptive method for GEP in predicting the compressive strength of concrete.

5- HCVCM improve the accuracy SBSR 2.6% in coefficient of determination, 6.5% in RMSE, 3.75% in NMSE, 11.4% in MAPE, 69.2% in maximum positive error, and 8.9% in the maximum negative error.

6- HCVCM-ANFIS is selected as the best model to predict the compressive strength of concrete, HCVCM improve the accuracy ANFIS 5% in coefficient of determination, 10% in RMSE, 3% in NMSE, 20% in MAPE, and 7% in the maximum negative error.

References

[1] Feng D, Liu Z, Wang X, Chen Y, Chang J, Wei D, et al. Machine learning-based compressive strength prediction for concrete : An adaptive boosting approach. Constr Build Mater 2020;230:117–28.

[2] Deng F, He Y, Zhou S, Yu Y, Cheng H, Wu X. Compressive strength prediction of recycled concrete based on deep learning. Constr Build Mater 2018;175:562–9. doi:10.1016/j.conbuildmat.2018.04.169.

[3] ASTM C1383-15, Standard Test Method for Measuring the P-Wave Speed and the Thickness of Concrete Plates Using the Impact-Echo Method, ASTM International, West Conshohocken, PA, 2015, www.astm.org n.d. doi:10.1520/C1383-15.

[4] ASTM C597-16, Standard Test Method for Pulse Velocity Through Concrete, ASTM International, West Conshohocken, PA, 2016, www.astm.org n.d. doi:10.1520/C0597-16.

[5] Abu Yaman M, Abd Elaty M, Taman M. Predicting the ingredients of self compacting concrete using artificial neural network. Alexandria Eng J 2017;56:523–32. doi:10.1016/j.aej.2017.04.007.

[6] Duan ZH, Kou SC, Poon CS. Prediction of compressive strength of recycled aggregate concrete using artificial neural networks. Constr Build Mater 2012;40:1200–6. doi:10.1016/j.conbuildmat.2012.04.063.

[7] Shishegaran A, Ghasemi MR, Varaee H. Performance of a novel bent-up bars system not

interacting with concrete. Front Struct Civ Eng 2019;13:1301–15.

[8]   Shishegaran A, Daneshpajoh F, Taghavizade H, Mirvalad S. Developing conductive concrete containing wire rope and steel powder wastes for route deicing. Constr Build Mater 2020;232:117184.

[9]   Na UJ, Park TW, Feng MQ, Chung L. Neuro-fuzzy application for concrete strength prediction using combined non-destructive tests. Mag Concr Res 2009;61:245–56. doi:10.1680/macr.2007.00127.

[10]  Poorarbabi A, Ghasemi M, Moghaddam MA. Concrete compressive strength prediction using non-destructive tests through response surface methodology. Ain Shams Eng J 2020. doi:10.1016/j.asej.2020.02.009.

[11]  Rashid K, Waqas R. Compressive strength evaluation by non-destructive techniques : An automated approach in construction industry. J Build Eng 2017;12:147–54. doi:10.1016/j.jobe.2017.05.010.

[12]  J. Helal, M. Sofi PM. Non-Destructive Testing of Concrete : A Review of Methods. Electron J Struct Eng 2015;14.

[13]  Mulik Nikhil V, Balkiminal R, Chhabria Deep S, Ghare Vijay D, Tele Vishal S. the use of combined non destructive testing in the concrete strenght assessment from laboratory specimens and existing buildings. Int J Curr Eng Sci Res 2015;2:55–9.

[14]  BS 1881: Part 203: 1986: Measurement of Velocity of Ultrasonic Pulses in Concrete, BSI, U.K., 1986. n.d.

[15]  ARIÖZ Ö, Tuncan A, Tuncan M, Kavas T, Ramyar K, KILINÇ K, et al. Use of Combined Non-Destructive Methods to Assess the Strength of Concrete in Structures. Afyon Kocatepe Üniversitesi Fen Ve Mühendislik Bilim Derg 2009;9:147–54.

[16]  Kewalramani MA, Gupta R. Concrete compressive strength prediction using ultrasonic pulse velocity through artificial neural networks. Autom Constr 2006;15:374–9. doi:10.1016/j.autcon.2005.07.003.

[17] Jain A, Kathuria A, Kumar A, Verma Y, Murari K. Combined use of non-destructive tests for assessment of strength of concrete in structure. Procedia Eng 2013;54:241–51.

[18] ASTM C805 / C805M-18, Standard Test Method for Rebound Number of Hardened Concrete, ASTM International, West Conshohocken, PA, 2018, www.astm.org n.d. doi:10.1520/C0805_C0805M-18.

[19] Jain A, Kathuria A, Kumar A, Verma Y, Murari K. Combined use of non-destructive tests for assessment of strength of concrete in structure. Procedia Eng 2013;54:241–51. doi:10.1016/j.proeng.2013.03.022.

[20] Huang Q, Gardoni P, Hurlebaus S. Predicting Concrete Compressive Strength Using Ultrasonic Pulse Velocity and Rebound Number 2012:2012.

[21] Cristofaro MT, Viti S, Tanganelli M. New predictive models to evaluate concrete compressive strength using the sonreb method. J Build Eng 2019:100962. doi:10.1016/j.jobe.2019.100962.

[22] Asteris PG, Ashrafian A, Rezaie-balf M. Prediction of the compressive strength of self-compacting concrete using surrogate models 2019;2:137–50.

[23] Topcu IB, Sarıdemir M. Prediction of compressive strength of concrete containing fly ash using artificial neural networks and fuzzy logic. Comput Mater Sci 2008;41:305–11.

[24] Bilgehan M, Turgut P. Artificial neural network approach to predict compressive strength of concrete through ultrasonic pulse velocity. Res Nondestruct Eval 2010;21:1–17.

[25] Ashrafian A, Shokri F, Javad M, Amiri T, Mundher Z. Compressive strength of Foamed Cellular Lightweight Concrete simulation : New development of hybrid artificial intelligence model. Constr Build Mater 2020;230:117048. doi:10.1016/j.conbuildmat.2019.117048.

[26] Albuthbahak O. Prediction of concrete compressive strength using supervised machine learning models through ultrasonic pulse velocity prediction of concrete compressive strength using supervised machine learning models through ultrasonic pulse velocity and mix parameter. Rom J Mater 2019;49:232–43.

[27] Ahmadi-Nedushan B. An optimized instance based learning algorithm for estimation of

compressive strength of concrete. Eng Appl Artif Intell 2012;25:1073–81. doi:10.1016/j.engappai.2012.01.012.

[28] Huang Q, Gardoni P, Hurlebaus S. Predicting concrete compressive strength using ultrasonic pulse velocity and rebound number. ACI Mater J 2011;108:403.

[29] Ongpeng J, Soberano M, Oreta A, Hirose S. Artificial neural network model using ultrasonic test results to predict compressive stress in concrete. Comput Concr 2017;19:59–68. doi:10.12989/cac.2017.19.1.059.

[30] Trtnik G, Kavčič F, Turk G. Prediction of concrete strength using ultrasonic pulse velocity and artificial neural networks. Ultrasonics 2009;49:53–60.

[31] Tenza-Abril AJ, Villacampa Y, Solak AM, Baeza-Brotons F. Prediction and sensitivity analysis of compressive strength in segregated lightweight concrete based on artificial neural network using ultrasonic pulse velocity. Constr Build Mater 2018;189:1173–83. doi:10.1016/j.conbuildmat.2018.09.096.

[32] Ashrafian A, Javad M, Amiri T, Rezaie-balf M, Ozbakkaloglu T. Prediction of compressive strength and ultrasonic pulse velocity of fiber reinforced concrete incorporating nano silica using heuristic regression methods. Constr Build Mater 2018;190:479–94. doi:10.1016/j.conbuildmat.2018.09.047.

[33] Topçu İB, Sarıdemir M. Prediction of mechanical properties of recycled aggregate concretes containing silica fume using artificial neural networks and fuzzy logic. Comput Mater Sci 2008;42:74–82.

[34] Topçu İB, Sarıdemir M. Prediction of rubberized concrete properties using artificial neural network and fuzzy logic. Constr Build Mater 2008;22:532–40.

[35] Mansouri I, Hu JW, Kisi O. Novel predictive model of the debonding strength for masonry members retrofitted with FRP. Appl Sci 2016;6:337.

[36] Madandoust R, Bungey JH, Ghavidel R. Prediction of the concrete compressive strength by means of core testing using GMDH-type neural network and ANFIS models. Comput Mater Sci

2012;51:261–72.

[37] Shishegaran A, Saeedi M, Kumar A, Ghiasinejad H. Prediction of air quality in Tehran by developing the nonlinear ensemble model. J Clean Prod 2020:120825.

[38] Domingo R, Hirose S. Correlation between concrete strength and combined nondestructive tests for concrete using high-early strength cement. Sixth Reg. Symp. Infrastruct. Dev., 2009, p. 12–3.

[39] Bahmani P. Concrete strength estimation by means of combined ultrasonic and Schmidt rebound hammer methods using nonlinear regression analysis and neural network. MSc. Thesis, Pooyandegan Danesh Institution of Higher Education, 2015.

[40] Asteris PG, Mokos VG. Concrete compressive strength using artificial neural networks. Neural Comput Appl 2019;2. doi:10.1007/s00521-019-04663-2.

[41] Chen Y, Wang Z, Liu Y, Zuo MJ, Huang H-Z. Parameters determination for adaptive bathtub-shaped curve using artificial fish swarm algorithm. Proc - Annu Reliab Maintainab Symp 2012.

[42] Asteris PG, Ashrafian A, Rezaie-balf M. Prediction of the compressive strength of self-compacting concrete using surrogate models Panagiotis. Comput Concr 2019;24:137–50. doi:10.12989/cac.2019.24.2.137.

[43] Shishegaran A, Reza M, Karami B, Rabczuk T. Computational predictions for estimating the maximum deflection of reinforced concrete panels subjected to the blast load. Int J Impact Eng 2020;139:103527. doi:10.1016/j.ijimpeng.2020.103527.

[44] Rutkowski L. FLEXIBLE NEURO-FUZZY SYSTEMS. Structures, Learning and Performance Evaluation. Czestochowa-Polonia 2004.

[45] Ahmadi-Nedushan B. Prediction of elastic modulus of normal and high strength concrete using ANFIS and optimal nonlinear regression models. Constr Build Mater 2012;36:665–73. doi:10.1016/j.conbuildmat.2012.06.002.

[46] Mirrashid M, Givehchi M, Miri M, Madandoust R, Branch Z, Faculty CE, et al. PERFORMANCE INVESTIGATION OF NEURO-FUZZY SYSTEM FOR EARTHQUAKE PREDICTION 2016;17:213–23.

[47]   Ghiasi R, Ghasemi MR, Noori M. Comparative studies of metamodeling and AI-Based techniques in damage detection of structures. Adv Eng Softw 2018;125:101–12.